\documentclass[lettersize,journal]{IEEEtran}

\usepackage{multirow}

\usepackage{cite}
\usepackage{amsmath,amssymb,amsfonts}
\usepackage{graphicx}
\usepackage{subcaption}
\usepackage{pgfplots}

\usepackage{xcolor}      
\usepackage{longtable}   
\usepackage{algorithm}
\usepackage{algorithmic}
\usepackage[hidelinks]{hyperref} 
\usepackage[capitalise]{cleveref} 

\title{Co-planning of Flight Corridors and Communication Infrastructure for Urban Drone Logistics Networks}

\author{Yingjie He, Yikang Wang, Zhenyu Gao}

\begin{document}

\maketitle

\begin{abstract}

Reliable wireless connectivity is essential for urban air mobility (UAM) networks in dense urban environments. It is therefore imperative to carefully plan the supporting communication infrastructure for UAM flight corridors. Most existing works optimize communication infrastructure and UAV flight paths independently, often leading to unnecessary base station (BS) deployment or excessive flight detours. This paper studies the joint optimization of BS deployment and UAV flight corridors in complex urban environments, aiming to minimize both infrastructure investment and flight distance while satisfying communication quality constraints. We propose CR-CMAB, a channel reciprocity-guided combinatorial multi-armed bandit framework. The framework constructs high-fidelity radio maps using 3D ray tracing, selects BS combinations via coverage-aware CMAB search, and dynamically expands the search space by identifying promising BS locations through channel reciprocity. Experimental results from a detailed case study demonstrate that CR-CMAB outperforms baseline methods with moderate computational time, yielding more strategically positioned BSs and shorter flight corridors. This study offers a practical planning perspective for cost-effective and communication-reliable UAM deployment in future smart cities.

\end{abstract}

\def\abstractname{Note to Practitioners}
\begin{abstract}
Communication systems are among the most prominent constraints for urban air mobility (UAM) operations. To deploy communication infrastructures for UAM, planners must balance two conflicting objectives in the process: minimizing cost while maximizing efficiency. System operators therefore have a strong interest in strategically planning communication infrastructure along UAM flight corridors. To better tackle this challenge, we propose the joint optimization of UAM base stations and flight corridors. Our proposed optimization framework leverages several rigorous components, including high-precision simulation, a combinatorial multi-armed bandit algorithm, and the principle of reciprocity, to collaboratively plan the locations of base stations and the configuration of a network of flight corridors. We validate our approach through a case study in a digital 3D urban environment, and the results show that our approach achieves the best efficiency in terms of total flight distance at each specific infrastructure cost level. We believe this work paves the way for further advancements leading to optimal design schemes for UAM communication infrastructure.
\end{abstract}


\section{Introduction}

Urban Air Mobility (UAM) is an emerging concept that enables efficient transportation within urban environment using unmanned aerial vehicles (UAVs) and electric vertical takeoff and landing (eVTOL) aircraft. By exploiting low-altitude urban airspace, UAM supports a broad range of applications, including passenger transit, cargo delivery, and urban monitoring, offering a promising solution to alleviate terrestrial traffic congestion and enhance the sustainability and resilience of smart cities \cite{wang2025uamnetwork, dissanayaka2024review,park2026urbananalytics}.

It is envisioned that UAM vehicles will operate within a service network, where vertiports and waypoints function as nodes and flight corridors function as links \cite{wei2024risk}. Flight corridors are designated virtual highways in the sky that aircraft are required to follow, offering navigation, reduced airspace complexity, and improved operational safety and efficiency~\cite{gao2024equitable}. However, maintaining safe and sustainable aerial vehicle operations within flight corridors creates substantial infrastructure challenges. Communication is an essential component of the necessary supporting infrastructure. Existing city-wide communication infrastructure falls short of meeting the connectivity requirements of an aerial vehicle network in urban low-altitude airspace. In addition, the dense concentration of high-rise buildings creates severe urban canyon effects, leading to critical line-of-sight (LoS) blockages and multipath fading \cite{gapeyenko2021line}. Such a heavily obstructed radio environment significantly degrades the reliability of wireless communication links~\cite{savkin2024joint}. A fundamental requirement for UAM safety is therefore the strategic deployment of reliable communication infrastructure, specifically base stations (BSs), to support networked flight corridor operations. Likewise, to prevent communication outages, flight corridors must be planned to strictly satisfy communication quality constraints, ensuring continuous radio coverage throughout the flight trajectories \cite{kim20226g, zaid2023evtol}.

Recent studies have primarily focused on either optimizing flight paths under fixed BS deployments or placing BSs for predetermined routes \cite{bo2024low, wang2021learning}. These decoupled strategies neglect the strong coupling between UAV mobility and radio coverage, often leaving flight paths trapped in communication blind spots or leading to inefficient use of BS resources. Hence, joint optimization of BS placement and flight corridors is essential to effectively balance the objectives of both UAM service providers and wireless network operators.

While a co-planning approach is theoretically ideal, realizing it in complex 3D urban environments imposes a high computational burden. The search space of candidate BS locations and 3D flight paths grows exponentially, rendering conventional combinatorial optimization algorithms highly inefficient and prone to suboptimal solutions. To address this problem, we propose a novel Channel Reciprocity-guided Combinatorial Multi-Armed Bandit (CR-CMAB) framework for this joint optimization task. The main contributions of this work are summarized as follows:
\begin{itemize}
    \item \textbf{Joint optimization approach within a digital twin environment.} We formulate the joint optimization of BS deployment and UAM flight corridors within a radio frequency simulation environment. By constructing high-fidelity radio maps, this environment captures complex urban radio propagation dynamics, establishing a physical foundation for the subsequent search algorithm.
    
    \item \textbf{Coverage-aware CMAB BS selection.} We formulate BS selection as a coverage-aware combinatorial multi-armed bandit (CMAB) problem, where each candidate BS is an arm and each selected subset forms a super arm. The selection strategy combines upper confidence bound (UCB) exploration with waypoint coverage rewards, encouraging complementary BS selections that jointly support the flight corridor while efficiently exploring the combinatorial space.
    
    \item \textbf{Reciprocity-driven dynamic space expansion.} We propose a dynamic action space expansion mechanism grounded in the principle of reciprocity. Using temporary UAV paths as radio frequency probes, this mechanism infers promising BS locations and injects them into the candidate pool. This feedback loop intelligently expands the search space to include higher-quality BS locations.
\end{itemize}

\section{Related Works}

\subsection{Base Station Deployment}

BS deployment is a fundamental problem that determines the number and locations of BS sites under coverage, capacity, quality-of-service (QoS), and other constraints \cite{chen2024modeling}. At its core, BS deployment can be viewed as a combinatorial optimization problem, since it requires selecting a subset of BSs from a large pool of candidate sites. In dense urban environments, the difficulty increases further because blockage makes the feasible deployment space tightly coupled with the surrounding built environment.

A major line of work formulates BS deployment as a combinatorial optimization problem. Such problems are commonly solved by discretizing the deployment region and searching over candidate BS locations using heuristic or metaheuristic algorithms. Recent explorations include swarm intelligence approaches such as grey wolf optimization for BS deployment in Internet-of-Things networks \cite{pliatsios2021drone}, the random-walk sparrow search algorithm for 5G private network planning \cite{chang20215g}, and the dynamically adjusted quantum genetic algorithm for 5G BS selection \cite{li2025optimization}. These heuristic methods are effective, but they often require extensive search and repeated communication quality evaluation in complex urban environments.

Learning-based methods have recently emerged as an alternative to the aforementioned metaheuristics. A related work uses reinforcement learning (RL) with digital network twins to learn deployment policies from pathloss predictions \cite{lee2025autobs}. Multi-objective deep reinforcement learning has also been used to place BSs while balancing coverage and localization requirements for urban transportation scenarios \cite{al2024multi}. However, learning-based methods often require substantial training time and may exhibit limited generalization when transferred across different scenarios.

Cost is a dominant factor in BS deployment. Wang \emph{et al.} adopted geographic information systems (GIS) and heuristic optimization to maximize service coverage while reducing construction and operation cost \cite{wang2020optimizing}. Prasad \emph{et al.} jointly optimized BS location, BS density, and transmit power to minimize network operational cost while satisfying user coverage constraints \cite{prasad2018joint}. Chiaraviglio \emph{et al.} formulated 5G network planning under electromagnetic field constraints as an NP-hard optimization problem, where the objective balances service quality and the installation cost of next-generation NodeBs (gNBs) \cite{chiaraviglio20215g}. Together, these works show that BS deployment is not only a coverage problem but also a cost-constrained planning problem.

For UAV services, BS deployment should be evaluated along the intended air corridor rather than only over a fixed planar service region. Lin \emph{et al.} proposed a low-altitude communication network planning method for air route coverage, where existing ground BSs are selected for upgrade to support aerial links along predefined routes \cite{lin2025low}. Kabashkin \emph{et al.} studied communication infrastructure design for UAV operations in air mobility corridors and analyzed repeater deployment strategies \cite{kabashkin2025communication}. These studies highlight the importance of corridor-aware communication supply for UAVs, but they still assume that the flight corridor is given in advance. As a result, the deployment strategy may become inefficient when the route passes through regions with poor communication coverage, leading to unnecessary additional BS installations.

\subsection{Communication-Aware UAV Path Planning}

Conventional UAV path planning problems mainly consider path efficiency, obstacle avoidance, and collision prevention, while ignoring communication conditions along the route \cite{meng2023learning}. However, insufficient communication support may degrade command and control (C2) link reliability and affect UAV operational safety. Communication-aware UAV path planning addresses this limitation by incorporating wireless link quality metrics (e.g., received signal strength, outage probability, and latency) into trajectory design. This allows the planned trajectory to maintain more reliable connectivity throughout the flight.

A common approach involves constructing a radio coverage map and planning the UAV trajectory over this communication-aware representation. Song \emph{et al.} used partial channel knowledge maps (CKMs) to optimize trajectories for cellular-connected UAVs in complex environments \cite{song2026trajectory}. Chen \emph{et al.} studied radio-map-based trajectory design for a UAV cargo pickup and delivery system, where the UAV route is optimized to satisfy communication requirements throughout the mission \cite{chen2024optimal}. Other works use cellular coverage maps to guide UAVs toward regions with better communication reliability, using graph search algorithms or RL to trade path length for link quality \cite{cao2025energy,duo2024joint}.

These studies show that communication constraints can significantly reshape UAV trajectories. However, they usually assume that the BS deployment or the radio map is already given. As a result, the UAV can only adapt its path within a fixed communication environment. When the existing BS deployment creates severe communication blind spots, path-only optimization may lead to long detours or even infeasible routes. This limitation highlights the need to jointly consider BS deployment and flight corridor design, rather than optimizing either component in isolation.

\subsection{The Research Gap: Joint Optimization of BS deployment and UAV Network Corridors}

BS deployment studies usually focus on selecting communication infrastructure for given service regions, while communication-aware UAV path planning studies typically optimize trajectories under a fixed radio environment. Such separate treatments may lead to suboptimal solutions, particularly for UAV network operations, since the selected BSs determine where reliable communication is available, while the flight corridor determines where communication service is actually required. Jointly optimizing BS deployment and flight corridor design can therefore reduce unnecessary infrastructure deployment and avoid excessive trajectory detours.

However, studies that explicitly couple these two components for urban UAV networks remain limited. Huang \emph{et al.} proposed a heterogeneous dual-network framework for emergency delivery UAVs, where a communication support network is coordinated with a delivery path network \cite{huang2026heterogeneous}. Although the framework considers both communication assurance and path planning, its solution is decomposed into sequential modules, where the path planning stage mainly adapts to a pre-established communication support network, rather than being optimized simultaneously with BS deployment. Li \emph{et al.} considered the concept of joint air corridor planning and BS deployment for low-altitude ISAC networks with a channel map \cite{li2025channel}. However, their work does not account for network operations, and the evaluation scenario remains relatively simplistic, overlooking the full propagation complexity of realistic urban environments.

To address these limitations, this paper studies joint BS deployment and networked UAV flight corridor planning in complex urban environments. The proposed framework builds accurate radio maps using ray tracing, selects BSs through a coverage-aware CMAB search, and leverages channel reciprocity to expand the BS search space by identifying new candidate locations with strong potential to fill corridor coverage gaps. It also evaluates infrastructure and trajectory objectives through interpretable BS deployment and UAV operation costs, enabling a more practical tradeoff between infrastructure expenditure and flight efficiency.

\section{Methodology}

\subsection{Proposed Framework}

\begin{figure*}[t]  
    \centering
    \includegraphics[width=\textwidth]{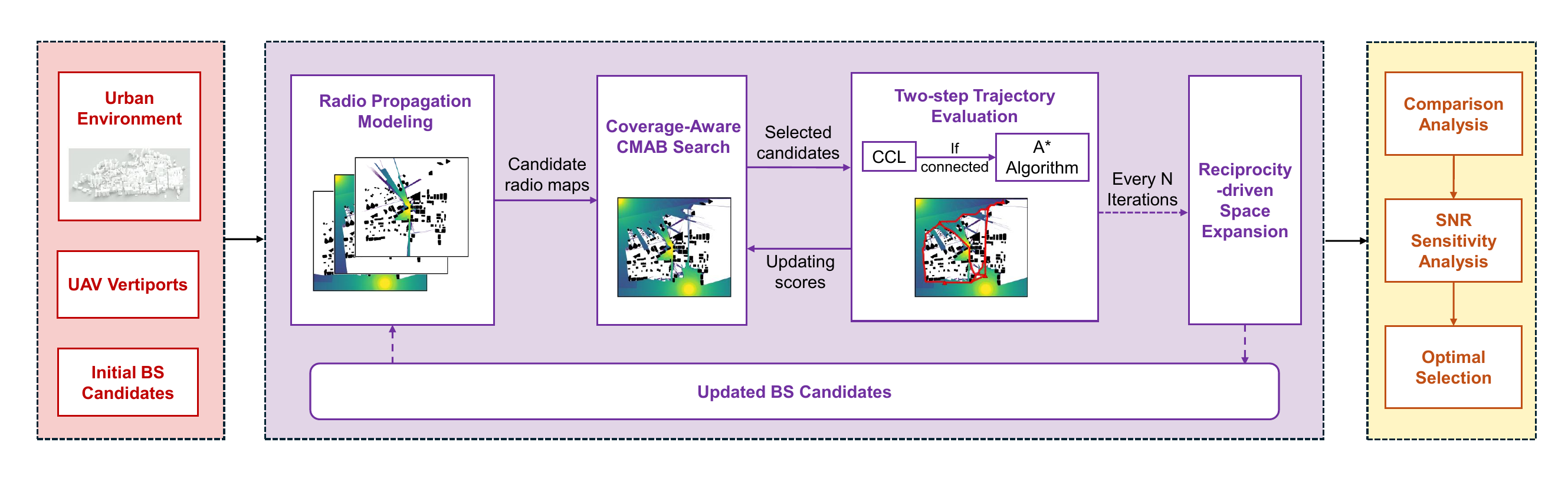} 
    \caption{The CR-CMAB framework for joint optimization of BS deployment and networked flight corridors.}
    \label{fig:framework}
\end{figure*}

We consider the joint optimization of flight corridors and communication infrastructure for a UAV network in a complex urban environment. The general form of this multi-objective optimization problem can be formulated as follows:

\begin{equation}
\begin{aligned}
\min_{\mathbf{z}, \pi} \quad & \bigg[ L(\pi), \; C(\mathbf{z}) \bigg] \\
\text{s.t.} \quad & h_{\text{obs}}(\pi) = 0, \\
& Q_{\mathbf{z}}(\mathbf{s}) \ge \gamma_{\text{th}}, \quad \forall \mathbf{s} \in \pi, \\
& \mathbf{z} \in \{0, 1\}^M, \; \pi \in \Pi,
\end{aligned}
\label{eq:problem}
\end{equation}
where the objective vector aims to simultaneously minimize the length of the total flight paths, denoted by $L(\pi)$, and the total infrastructure cost of the deployed base stations, $C(\mathbf{z})$. Here, $\mathbf{z} \in \{0, 1\}^M$ is a binary decision vector indicating the deployment status across $M$ candidate base station locations. $\pi$ denotes the set of networked UAV flight paths confined to a designated cruising altitude, selected from $\Pi$, which represents the universal set of all feasible paths. The optimal joint solution $(\mathbf{z}^*, \pi^*)$ yielding the Pareto frontier must strictly adhere to both physical and communication boundaries. Specifically, the physical constraint $h_{\text{obs}}(\pi)$ prevents the UAV from colliding with surrounding buildings. The connectivity constraint dictates that the spatial signal quality, $Q_{\mathbf{z}}(\mathbf{s})$, must satisfy or exceed the minimum threshold $\gamma_{\text{th}}$ for every location $\mathbf{s}$ along $\pi$. This guarantees that the UAV maintains a continuous and reliable radio link throughout the entire flight corridor.

The challenge of this co-planning problem lies in the exponential growth of the search space. The combinatorial complexity of the problem is further exacerbated by the communication constraints: in dense urban scenarios, calculating the exact signal quality $Q_{\mathbf{z}}(\mathbf{s})$ for every $(\mathbf{z}, \mathbf{s})$ pair requires radio propagation assessments, which creates a major computational bottleneck.

To tackle this problem, we propose CR-CMAB, a novel framework that integrates coverage-aware CMAB search with reciprocity-driven space expansion, as illustrated in Fig. \ref{fig:framework}. The core workflow consists of four modules. First, the ray tracing method generates the radio maps of candidate BSs. Then, the coverage-aware CMAB module selects a BS combination under the given budget. The selected candidates are evaluated by the two-step trajectory evaluation module, where the connected component labeling (CCL) checks corridor connectivity and A* computes the feasible path. The evaluation result is fed back to update the CMAB scores. After every fixed number of iterations, reciprocity-driven space expansion adds new promising BS candidates to the pool, allowing the search space to be refined progressively.

\subsection{Radio Propagation Modeling via Ray Tracing}

To quantify the communication capability of each candidate BS, it is necessary to construct high-precision radio maps, which are essentially spatial heatmaps representing the geographical distribution of the signal-to-noise (SNR) ratio. By capturing the electromagnetic propagation characteristics over the urban topology, radio maps provide a deterministic communication metric for any potential flight.

Existing methods for constructing radio maps generally fall into three categories: empirical models, data-driven methods, and ray tracing. Empirical models, such as the log-distance path loss model (LDPL), are computationally efficient but fail to capture blockage and multipath effects in realistic propagation environments. On the other hand, emerging data-driven methods (e.g., RadioUNet \cite{levie2021radiounet} and PMNet \cite{lee2023pmnet}) require extensive training data and often show limited robustness across different urban layouts. Consequently, we adopt a deterministic 3D ray tracing approach. Based on geometric optics, this method captures complex electromagnetic propagation mechanisms, including penetrations, reflections, and diffractions, providing high-fidelity radio maps without the need for prior training datasets.

Let $\mathbf{b}_i$ denote the location of the $i$-th candidate BS, and let $\mathbf{s}=(x,y)$ represent an arbitrary coordinate on the horizontal flight plane. The channel propagation gain from $\mathbf{b}_i$ to $\mathbf{s}$, calculated by ray tracing, is denoted by $L_{\text{ch}}(\mathbf{b}_i,\mathbf{s})$. To construct the SNR radio maps defined previously, the received signal power at location $\mathbf{s}$ from the candidate BS at $\mathbf{b}_i$ is calculated using the link budget equation:
\begin{equation} \label{eq:rx_power}
P(\mathbf{b}_i,\mathbf{s}) = P_t + G_t + L_{\text{ch}}(\mathbf{b}_i,\mathbf{s}) + G_r,
\end{equation}
where $P_t$ is the BS transmitter output power, and $G_t$ and $G_r$ are the transmitter and receiver antenna gains, respectively. Then, the noise power $P_n$ at the receiver is modeled as:
\begin{equation} \label{eq:noise_power}
P_n = N_0 + 10\log_{10}(B) + N_F,
\end{equation}
where $N_0$ represents the thermal noise power spectral density, $B$ is the channel bandwidth, and $N_F$ is the hardware noise figure. With both the received signal power and noise derived, the single-BS SNR at $\mathbf{s}$ is formulated as:
\begin{equation} \label{eq:snr_db}
\begin{split}
q_i(\mathbf{s}) &= P(\mathbf{b}_i,\mathbf{s}) - P_n\\
&= P_t + G_t + G_r + L_{\text{ch}}(\mathbf{b}_i,\mathbf{s}) - P_n.
\end{split}
\end{equation}

When multiple BSs exist in the same airspace, we assume that the UAV maintains an active connection with only the BS providing the strongest signal. Consequently, the complete SNR radio map under the BS decision vector $\mathbf{z}$, denoted as $Q_{\mathbf{z}}(\mathbf{s})$, is formulated as:
\begin{equation} \label{eq:joint_snr}
Q_{\mathbf{z}}(\mathbf{s}) = \max_{i \in \{1, 2, \dots, M\}, z_i=1} q_i(\mathbf{s}), \quad \forall \mathbf{s}.
\end{equation}

\subsection{Coverage-Aware Combinatorial Multi-Armed Bandit Search}

Given the candidate BS pool of size $M$, selecting $K$ BSs leads to `$M$ choose $K$' possible deployment combinations. This creates a massive combinatorial search space. In addition, evaluating the quality of each combination requires trajectory evaluation under communication constraints, which is relatively time-consuming. Consequently, directly assessing all candidate combinations is computationally prohibitive in realistic urban scenarios.

To reduce the search cost, we recast the BS selection process as a CMAB problem. Unlike a conventional bandit setting that selects a single arm at each decision step, CMAB selects a group of arms jointly, which is commonly referred to as a super arm. This formulation is suitable for BS deployment because the performance of one BS cannot be fully assessed in isolation. Instead, its value depends on how it cooperates with other selected BSs to provide continuous coverage along the flight corridor. Each candidate BS is regarded as an arm, and a super arm corresponds to a feasible BS deployment vector $\mathbf{z}$ defined in Eq.~\eqref{eq:problem}. For a given BS budget $K$, the selected super arm further satisfies
\begin{equation}
\sum_{i=1}^{M} z_i = K.
\end{equation}

To evaluate each arm during the search, we adopt the upper confidence bound (UCB) criterion. The UCB score consists of an empirical reward term and a confidence term, enabling a balance between exploitation and exploration.
 For candidate BS $i$, it is written as
\begin{equation}
UCB_i = \bar{r}_i + u_i.
\end{equation}

During the search, the algorithm maintains the number of times the $i$-th BS has been selected, denoted by $n_i$, and its accumulated normalized reward, denoted by $v_i$. The empirical reward is computed as
\begin{equation}
\bar{r}_i = \frac{v_i}{n_i+\epsilon},
\end{equation}
where $\epsilon$ is a small positive constant for numerical stability. A larger $\bar{r}_i$ indicates that $i$-th BS has more frequently appeared in high quality deployment combinations.

The confidence term $u_i$ is defined as
\begin{equation}
u_i = c\sqrt{\frac{\ln(\tau+1)}{n_i+\epsilon}},
\end{equation}
where $c$ controls the exploration strength and $\tau$ is the number of evaluated BS combinations. This term assigns a larger bonus to candidates with fewer selections, allowing less explored BSs to remain competitive during the search.

Although the UCB score captures the historical performance and uncertainty of each individual BS, it does not explicitly measure the performance complementarity among selected BSs. To promote such complementary effect, we introduce a waypoint-based coverage award. At the beginning of the search, the waypoint set can be empty. Once promising trajectories have been obtained from previous evaluations, we sample a set of corridor waypoints denoted by
\begin{equation}
\mathcal{W}=\{\mathbf{s}_1,\mathbf{s}_2,\dots,\mathbf{s}_{N_W}\}.
\end{equation}

For candidate BS $i$, its covered waypoint index set is defined as
\begin{equation}
\mathcal{C}_i =
\left\{
j \mid q_i(\mathbf{s}_j) \ge \gamma_{\text{th}},\;
\mathbf{s}_j \in \mathcal{W}
\right\}.
\end{equation}

During the greedy construction of one BS combination, let $\mathcal{U}$ denote the index set of waypoints that are still uncovered by the selected BSs. The marginal waypoint coverage gain of BS $i$ is
\begin{equation}
g_i =
\begin{cases}
\frac{|\mathcal{C}_i \cap \mathcal{U}|}{|\mathcal{W}|}, & |\mathcal{W}|>0,\\
0, & |\mathcal{W}|=0.
\end{cases}
\end{equation}
which gives a higher score to candidates that can cover currently uncovered parts of the sampled flight corridor, making the selected set of BSs less redundant in coverage.

The final selection score of candidate BS $i$ is
\begin{equation}
\label{eq:cmab_score}
\Phi_i =
\bar{r}_i
+
u_i
+
\lambda_c g_i,
\end{equation}
where $\lambda_c$ controls the influence of the waypoint coverage award.

At each iteration $t$, the algorithm greedily constructs one selected set $\mathcal{I}$. Starting from $\mathcal{I}=\emptyset$, it repeatedly chooses the candidate BS with the largest score $\Phi_i$. After a BS is selected, the uncovered waypoint set $\mathcal{U}$ is updated and the scores $\Phi_i$ of the remaining candidates are recomputed. To reduce redundant deployments in nearby locations, candidates whose physical distance to the selected BS is smaller than $d_{\min}$ are removed from the current action pool. Once $K$ BSs are selected, the corresponding deployment vector $\mathbf{z}$ is passed to the trajectory evaluation module to obtain the normalized reward $\rho^{(t)}$ for updating $v_i$. The updated statistics are then used to guide the construction of the next super arm.

\begin{algorithm}[t]
\caption{Coverage-Aware Combinatorial Multi-Armed Bandit Search}
\label{alg:cmab_search}
\begin{algorithmic}[1]
\REQUIRE Candidate BS location set $\mathcal{B}=\{\mathbf{b}_1,\mathbf{b}_2,\dots,\mathbf{b}_M\}$, waypoint set $\mathcal{W}$, BS budget $K$, threshold $\gamma_{\text{th}}$, exploration coefficient $c$, weight $\lambda_c$, minimum separation $d_{\min}$, maximum iterations $T$

\STATE Initialize $n_i\leftarrow 0$ and $v_i\leftarrow 0$ for all $\mathbf{b}_i\in\mathcal{B}$
\STATE Initialize $\tau\leftarrow 0$, $\rho^*\leftarrow -\infty$, $\mathbf{z}^*\leftarrow \mathbf{0}$, and cache $\mathcal{H}$
\STATE Compute $\mathcal{C}_i$ for all $\mathbf{b}_i\in\mathcal{B}$ if $\mathcal{W}\neq\emptyset$

\FOR{$t=1$ to $T$}
    \STATE $\mathcal{I}\leftarrow \emptyset$, $\mathcal{A}\leftarrow \mathcal{B}$
    \STATE $\mathcal{U}\leftarrow \{1,2,\dots,|\mathcal{W}|\}$
    
    \WHILE{$|\mathcal{I}|<K$ and $\mathcal{A}\neq\emptyset$}
        \FOR{each $\mathbf{b}_i\in\mathcal{A}$}
            \STATE $\bar{r}_i \leftarrow v_i/(n_i+\epsilon)$
            \STATE $u_i \leftarrow c\sqrt{\ln(\tau+1)/(n_i+\epsilon)}$
            \STATE Compute $g_i$ using $\mathcal{C}_i$ and $\mathcal{U}$
            \STATE $\Phi_i \leftarrow \bar{r}_i + u_i + \lambda_c g_i$
        \ENDFOR
        
        \STATE $\mathbf{b}_{i^*} \leftarrow \arg\max_{\mathbf{b}_i\in\mathcal{A}}\Phi_i$
        \STATE $\mathcal{I}\leftarrow \mathcal{I}\cup\{i^*\}$
        \STATE $\mathcal{U}\leftarrow \mathcal{U}\setminus \mathcal{C}_{i^*}$
        \STATE $\mathcal{A}\leftarrow \{\mathbf{b}_i\in\mathcal{A}: i\neq i^*, \|\mathbf{b}_i-\mathbf{b}_{i^*}\|_2\ge d_{\min}\}$
    \ENDWHILE
    
    \STATE Construct $\mathbf{z}^{(t)}$ from $\mathcal{I}$
    
    \IF{$\mathbf{z}^{(t)}$ is not in $\mathcal{H}$}
        \STATE Evaluate $\mathbf{z}^{(t)}$ and obtain normalized reward $\rho^{(t)}$
        \STATE Store $\rho^{(t)}$ in $\mathcal{H}$
        \STATE $\tau\leftarrow \tau+1$
    \ELSE
        \STATE Retrieve $\rho^{(t)}$ from $\mathcal{H}$
    \ENDIF
    
    \FOR{each $i\in\mathcal{I}$}
        \STATE $n_i\leftarrow n_i+1$
        \STATE $v_i\leftarrow v_i+\rho^{(t)}$
    \ENDFOR
    
    \IF{$\rho^{(t)}>\rho^*$}
        \STATE $\rho^*\leftarrow \rho^{(t)}$, $\mathbf{z}^*\leftarrow \mathbf{z}^{(t)}$
    \ENDIF
\ENDFOR

\RETURN $\mathbf{z}^*$
\end{algorithmic}
\end{algorithm}

\subsection{Two-Step Trajectory Evaluation}
Repeatedly performing communication-aware pathfinding often imposes a substantial computational burden, particularly within environments densely populated with obstacles. We implement a two-step trajectory evaluation strategy that segregates the preliminary assessment of topological connectivity from the subsequent derivation of the exact trajectory.

\subsubsection{Rapid pruning via CCL}
The first step serves as a computationally lightweight filter to swiftly eliminate invalid BS combinations failing to provide continuous end-to-end radio coverage. Given a specific deployment decision $\mathbf{z}$, the SNR radio map $Q_{\mathbf{z}}(\mathbf{s})$ is derived via Eq. \eqref{eq:joint_snr}. Subsequently, a binary communication coverage mask, denoted as $\mathcal{M}(\mathbf{s})$, is constructed by applying the predefined SNR threshold $\gamma_{\text{th}}$:
\begin{equation}
\mathcal{M}(\mathbf{s})
 = 
\begin{cases} 
1, & \text{if } Q_{\mathbf{z}}(\mathbf{s}) \ge \gamma_{\text{th}} \\ 
0, & \text{otherwise} 
\end{cases}
\end{equation}

Rather than directly applying trajectory search on $\mathcal{M}(\mathbf{s})$, we exploit the CCL algorithm from image processing and graph theory. By scanning the binary mask, the CCL algorithm partitions the feasible communication space into a set of disjoint, maximally connected sub-graphs. 

We then verify whether the vertiports (UAV's vertical takeoff and landing locations, the nodes in the network) reside within the same connected component. If they are distributed across disconnected regions, it indicates the absence of a continuous communication corridor. Consequently, the trajectory evaluation is immediately terminated, and a severe penalty is assigned to the combination $\mathbf{z}$. This rapid pruning mechanism circumvents the redundant computational overhead of exact pathfinding for disconnected BS deployments.

\subsubsection{Exact routing via communication-aware $A^*$}
Passing the CCL verification indicates that the vertiports are connected within the region satisfying the wireless connectivity constraint. Next, a communication-aware $A^*$ algorithm is employed to determine the optimal flight trajectory $\pi^*$. This trajectory operates within the feasible communication space ($\mathcal{M}(\mathbf{s}) = 1$) while safely avoiding physical building obstacles ($h_{\text{obs}}(\pi)=0$). 

The shortest total flight corridor length $L(\pi^*)$ is then computed and converted into a reward signal. For the deployment vector evaluated at iteration $t$, the raw reward is defined as
\begin{equation}
R^{(t)}
=
\begin{cases}
R_0 - L(\pi^*), & \text{if a feasible trajectory is found},\\
0, & \text{otherwise},
\end{cases}
\end{equation}
where $R_0$ is the base reward. A shorter feasible corridor therefore leads to a larger raw reward. To make the reward scale suitable for updating the bandit statistics in Algorithm~\ref{alg:cmab_search}, the raw reward is further normalized as
\begin{equation}
\rho^{(t)}
=
\min\left\{
1,
\max\left\{
0,
\frac{R^{(t)}-R_{\min}}{R_{\max}-R_{\min}}
\right\}
\right\},
\end{equation}
where $R_{\min}$ and $R_{\max}$ denote the lower and upper normalization bounds, respectively. The normalized reward $\rho^{(t)}$ is then used to update the accumulated reward $v_i$ of each selected BS in Algorithm~\ref{alg:cmab_search}.

\subsection{Reciprocity-Driven Space Expansion}

The CMAB search operates on a finite candidate BS pool. Although this pool makes the combinatorial search tractable, it also imposes a structural limitation on the final solution. In dense urban environments, a small number of geometrically sensitive locations may determine whether a corridor can maintain continuous radio coverage. If the initial candidate pool does not include these ``sweet spot'' locations, keeping the pool fixed throughout the search may prevent the algorithm from finding high-quality BS deployment layouts.

A direct way to enlarge the candidate pool is to sample additional BS locations and compute their radio maps. However, such a strategy is inefficient because most newly sampled positions may have little relevance to the current corridor structure. Instead of expanding the search space blindly, we use the paths discovered during the search to infer where new BS candidates should be introduced. The key idea is that a feasible or near-feasible flight corridor already contains information about communication demand. If a new BS location can provide coverage to multiple waypoints along this corridor, it is more likely to improve the subsequent CMAB search than an arbitrary location sampled from the whole map.

\begin{figure}[t]
    \centering
    \includegraphics[width=0.45\textwidth]{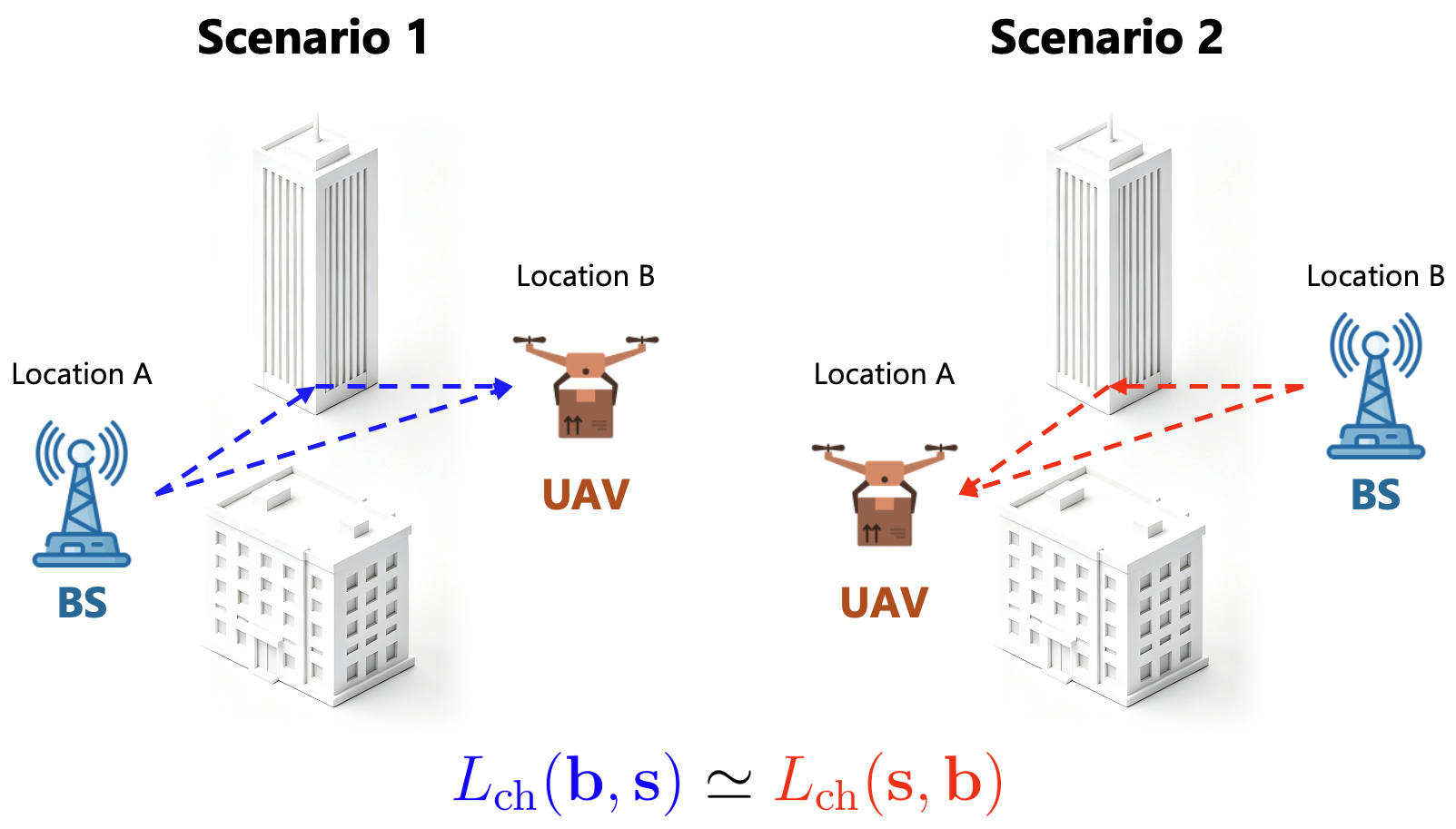}
    \caption{Illustration of channel reciprocity between a BS and a UAV waypoint in a complex urban environment. With swapped positions, the BS-to-UAV and UAV-to-BS links experience approximately the same propagation paths under a static reciprocal channel, including both LoS and reflected components.}
    \label{fig:reciprocity}
\end{figure}

The proposed mechanism is built upon the {\it reciprocity principle} of wave propagation, which is a statement of equivalence or symmetry~\cite{gao2023terrain}. In a static propagation environment, the transmitter and receiver sites may be interchanged without changing the amplitude or phase of the observed signal~\cite{chew2008reciprocity}. As shown in Fig.~\ref{fig:reciprocity}, for a BS deployment location $\mathbf{b}$ and a UAV waypoint $\mathbf{s}$, this property can be expressed as
\begin{equation}
L_{\text{ch}}(\mathbf{b},\mathbf{s}) \simeq L_{\text{ch}}(\mathbf{s},\mathbf{b}),
\end{equation}
where $L_{\text{ch}}(\mathbf{b},\mathbf{s})$ denotes the propagation gain from $\mathbf{b}$ to $\mathbf{s}$, and $L_{\text{ch}}(\mathbf{s},\mathbf{b})$ denotes the reverse link gain. This relation allows the algorithm to use UAV waypoints as temporary radio transmitters. Rather than testing many potential BSs in the forward direction, the algorithm transmits virtually from a small number of informative trajectory waypoints and observes which deployment locations can receive strong reciprocal signals. These locations are likely to be useful BS candidates because the reverse link suggests favorable forward coverage toward the same waypoints.

The waypoint set $\mathcal{W}$ used for the coverage award in Algorithm~\ref{alg:cmab_search} is sampled from promising trajectories obtained in recent CMAB iterations and is reused here for reciprocity-driven expansion. For each waypoint $\mathbf{s}_{\ell}\in\mathcal{W}$, a reciprocal SNR field is computed over the feasible BS deployment region, denoted by $\Omega_{\text{BS}}$. We use $\widetilde{q}_{\ell}(\mathbf{b})$ to represent the reciprocal SNR map generated by waypoint $\mathbf{s}_{\ell}$, where $\mathbf{s}_{\ell}$ acts as the temporary transmitter and $\mathbf{b}\in\Omega_{\text{BS}}$ is the receiving deployment location. The reciprocal coverage score of location $\mathbf{b}$ is defined as
\begin{equation}
G(\mathbf{b}) =
\sum_{\ell=1}^{N_W}
\mathbb{I}
\left\{
\widetilde{q}_{\ell}(\mathbf{b})
\ge
\gamma_{\text{th}}
\right\},
\quad \mathbf{b}\in\Omega_{\text{BS}} .
\end{equation}

The score $G(\mathbf{b})$ counts how many sampled corridor waypoints can be reciprocally covered by location $\mathbf{b}$. A larger value indicates that $\mathbf{b}$ is more relevant to the current corridor geometry. 

A location with a larger $G(\mathbf{b})$ value can reciprocally cover more sampled corridor waypoints, and is therefore assigned a higher priority during candidate generation. At the same time, candidate diversity is required because several neighboring locations may have similar reciprocal coverage scores. Selecting all of them would provide limited additional information to the CMAB search. Therefore, the new candidate set is selected by jointly considering the reciprocal coverage score and the spatial distribution among candidates.

Let $\mathcal{B}_{\text{new}}$ denote the set of newly injected BS candidates. It is given by
\begin{equation}
\mathcal{B}_{\text{new}}
=
\{\mathbf{b}_{M+1},\mathbf{b}_{M+2},\dots,\mathbf{b}_{M+M_{\text{new}}}\},
\quad
\mathbf{b}_{i}\in\Omega_{\text{BS}},
\end{equation}
where $M_{\text{new}}$ is the number of newly generated candidates. The selected locations are expected to have high reciprocal coverage scores while remaining spatially separated from each other. This design avoids repeatedly adding candidates from the same local region and increases the chance that different parts of the sampled corridor are supported by different new BS candidates.

Each location in $\mathcal{B}_{\text{new}}$ is then treated as a standard BS candidate. For every $\mathbf{b}_i\in\mathcal{B}_{\text{new}}$, a forward SNR radio map $q_i(\mathbf{s})$ is computed over the UAV flight plane using the radio propagation model introduced previously. These new radio maps are appended to the existing radio map library. The candidate pool is expanded from $M$ to $M+M_{\text{new}}$, so the following CMAB iterations can select both the original BS candidates and the reciprocity guided candidates.

\begin{algorithm}[t]
\caption{Reciprocity-Driven Space Expansion}
\label{alg:reciprocity_expansion}
\begin{algorithmic}[1]
\REQUIRE Promising trajectory set $\Pi_{\text{loc}}$, feasible BS deployment region $\Omega_{\text{BS}}$, current candidate pool $\mathcal{B}$, threshold $\gamma_{\text{th}}$, number of new candidates $M_{\text{new}}$

\STATE Sample waypoint set $\mathcal{W}=\{\mathbf{s}_{\ell}\}_{\ell=1}^{N_W}$ from $\Pi_{\text{loc}}$

\IF{$N_W=0$}
    \RETURN $\mathcal{B}$
\ENDIF

\FOR{$\ell=1$ to $N_W$}
    \STATE Compute $\widetilde{q}_{\ell}(\mathbf{b})$, $\forall \mathbf{b}\in\Omega_{\text{BS}}$
\ENDFOR

\FOR{each $\mathbf{b}\in\Omega_{\text{BS}}$}
    \STATE $G(\mathbf{b})\leftarrow
    \sum_{\ell=1}^{N_W}
    \mathbb{I}
    \left\{
    \widetilde{q}_{\ell}(\mathbf{b})
    \ge
    \gamma_{\text{th}}
    \right\}$
\ENDFOR

\STATE Initialize $\mathcal{B}_{\text{new}}\leftarrow\emptyset$

\WHILE{$|\mathcal{B}_{\text{new}}|<M_{\text{new}}$}
    \STATE $\mathbf{b}^*
    \leftarrow
    \arg\max_{\mathbf{b}\in\Omega_{\text{BS}}}
    G(\mathbf{b})$
    \STATE $\mathcal{B}_{\text{new}}\leftarrow
    \mathcal{B}_{\text{new}}\cup\{\mathbf{b}^*\}$
    \STATE Remove locations near $\mathbf{b}^*$ from $\Omega_{\text{BS}}$
\ENDWHILE

\FOR{each $\mathbf{b}_i\in\mathcal{B}_{\text{new}}$}
    \STATE Compute forward radio map $q_i(\mathbf{s})$, $\forall \mathbf{s}$
    \STATE Initialize its coverage set $\mathcal{C}_{i}$ and $n_i$, $v_i$
\ENDFOR

\STATE $\mathcal{B}\leftarrow\mathcal{B}\cup\mathcal{B}_{\text{new}}$
\STATE $M\leftarrow M+M_{\text{new}}$

\RETURN $\mathcal{B}$
\end{algorithmic}
\end{algorithm}

\section{Experiments}

\subsection{Experimental Setup}

\begin{figure}[t!]
    \centering
    \includegraphics[width=0.85\linewidth]{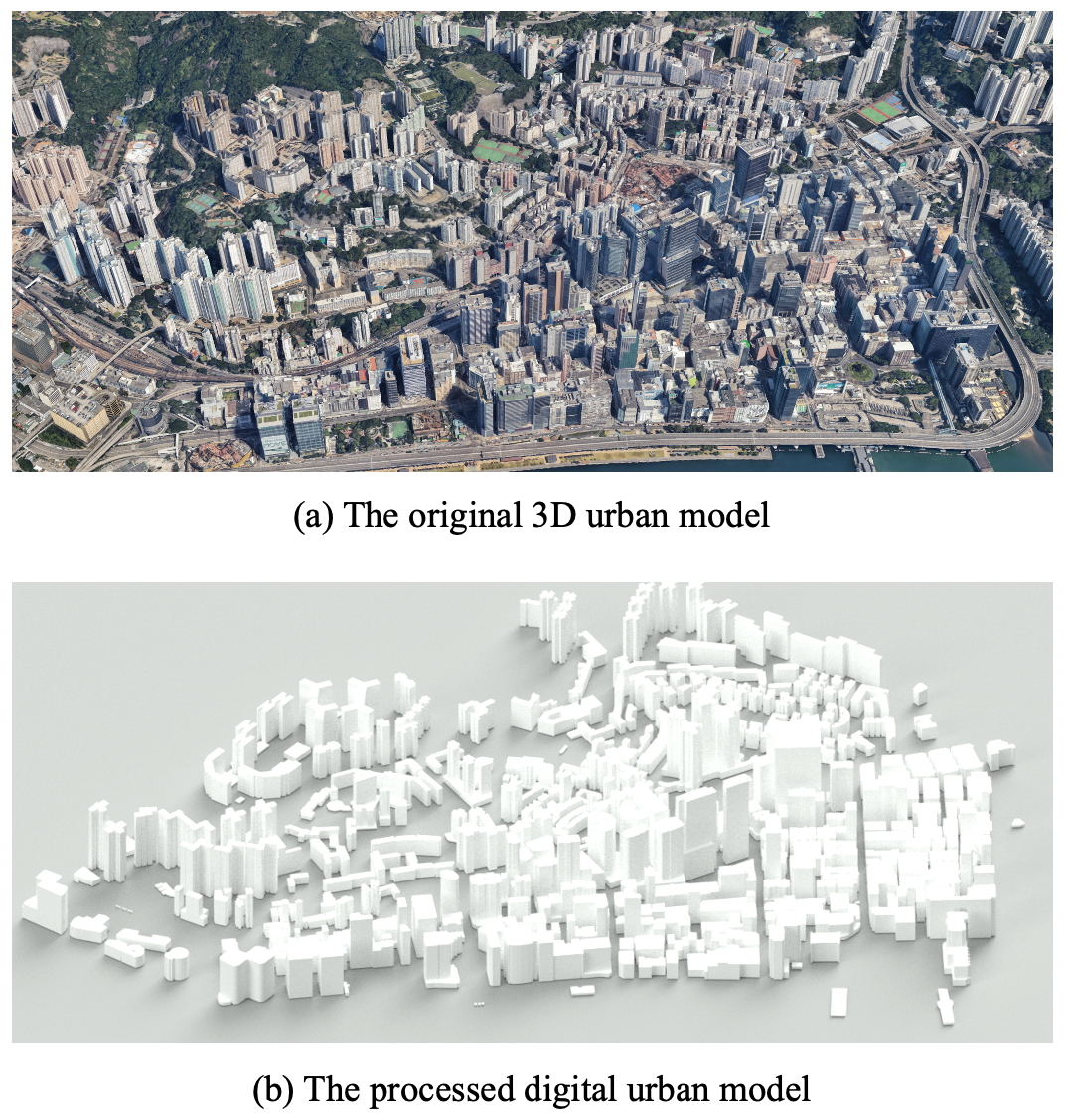}
    \caption{The 3D urban model used in the case study.}
    \label{fig:landsd_archviz}
\end{figure}

The experiments are conducted in a digital-twin urban scenario reconstructed from OpenStreetMap data. 
We select Kwun Tong, Hong Kong, as the case study area, where a UAV logistics network is particularly useful for delivering cargo of various natures. This typical industrial district represents a high-density urban environment with compact building blocks, irregular street layouts, and pronounced urban canyon characteristics.
These features provide a challenging testbed for co-planning BS deployment and UAV flight corridors.
In our numerical experiments, the UAV logistics network in Kwun Tong comprises four vertiport locations, forming six unique origin-destination flight corridors.
The digital 3D urban model used in the experiments is shown in Fig.~\ref{fig:landsd_archviz}.

For the UAV logistics network, the cruising altitude of air corridors is set to 50 m above ground level, which is a typical altitude for drone delivery under the 120 m constraint. The BSs serving the network are placed at a height of 10 m above ground level, installed on lampposts or building rooftops.

\begin{table}[t]
\centering
\caption{Main simulation parameters.}
\label{tab:simulation_parameters}
\normalsize
\renewcommand{\arraystretch}{1.08}
\setlength{\tabcolsep}{3.5pt}
\begin{tabular}{lc}
\hline
\textbf{Parameter} & \textbf{Value} \\
\hline
\multicolumn{2}{l}{\textit{\textbf{Communication parameters}}} \\
BS transmit power $P_t$ & $40~\mathrm{dBm}$ \\
BS antenna gain $G_t$ & $20~\mathrm{dB}$ \\
Carrier frequency $f_c$ & $39~\mathrm{GHz}$ \\
Bandwidth $B$ & $4\times10^8~\mathrm{Hz}$ \\
Noise figure $N_F$ & $5~\mathrm{dB}$ \\
UAV antenna gain $G_r$ & $8~\mathrm{dB}$ \\
SNR threshold $\gamma_{\mathrm{th}}$ & $20~\mathrm{dB}$ \\
Maximum ray depth & 3 \\
\hline
\multicolumn{2}{l}{\textit{\textbf{Scenario parameters}}} \\
Area size & $1600~\mathrm{m}\times1400~\mathrm{m}$ \\
Flight altitude & $50~\mathrm{m}$ \\
BS height & $10~\mathrm{m}$ \\
Number of vertiports & $4$ \\
\hline
\multicolumn{2}{l}{\textit{\textbf{Algorithm parameters}}} \\
Confidence term $c$ & $2$ \\
Coverage term $\lambda_c$ & $3$ \\
Minimum separation $d_{\min}$ & $100~\mathrm{m}$ \\
Candidate pool size $M$ & $300$ \\
Maximum iterations $T$ & $12000$ \\
Expansion interval $N$ & $3200$ iterations \\
New candidates number $M_{\mathrm{new}}$ & $20$ \\
\hline
\end{tabular}
\end{table}

NVIDIA Sionna RT~\cite{sionna} is used as the ray-tracing simulation platform for radio propagation modeling. 
Built upon GPU-accelerated physical ray tracing, Sionna RT provides an efficient simulation environment for constructing high-fidelity radio maps in complex urban environments. The main simulation parameters are summarized in Table~\ref{tab:simulation_parameters}.
All simulations are executed on a workstation equipped with an AMD Ryzen 9 9950X CPU, an NVIDIA GeForce RTX 5090 D GPU with CUDA 12.9 support, and 64~GiB running memory.

\subsection{Baseline Algorithms and Evaluation Metrics}

We compare the performance of our proposed method CR-CMAB with the following five baseline methods:

\begin{enumerate}
    \item \textbf{Greedy Coverage:} This baseline follows a conventional cellular deployment criterion that aims to maximize the covered area. It selects BSs according to their marginal coverage gain over the full radio map, with limited randomness introduced by a restricted candidate list.
    
    \item \textbf{Genetic Algorithm (GA):} GA evolves a population of BS combinations through selection, crossover, mutation, and elitism. The trajectory length returned by the evaluation module is used as the fitness value.

    \item \textbf{Particle Swarm Optimization (PSO):} PSO uses a continuous position vector to represent each particle. At each iteration, the top $K$ entries are mapped to a BS combination, and the particles are updated according to their personal and global best records.

    \item \textbf{Monte Carlo Tree Search (MCTS):} MCTS constructs a BS combination sequentially. Each node represents a partial selection, and the tree is updated through selection, expansion, rollout, and backpropagation.

    \item \textbf{CMAB Only:} This method removes reciprocity-driven space expansion from CR-CMAB while keeping the coverage-aware CMAB search. The candidate BS pool therefore remains fixed during the search.
\end{enumerate}

The flight corridor planning performance is first evaluated using the {\bf path length offset} with respect to the theoretical total shortest path length of the network without communication constraints:
\begin{equation}
\label{eq:path_offset}
\Delta L = L(\pi)-L(\pi_\mathrm{ref}),
\end{equation}
where $\pi$ denotes the communication-aware total path length and $\pi_\mathrm{ref}$ is the shortest path length through A* when only physical obstacles are considered. In the Kwun Tong UAV network, the reference path length is $L(\pi_\mathrm{ref})=6277.9~\mathrm{m}$.

\subsection{Comparison Analysis of Different Algorithms and Varying BS Budgets}

\begin{figure*}[h!]  
    \centering
    \includegraphics[width=0.925\textwidth]{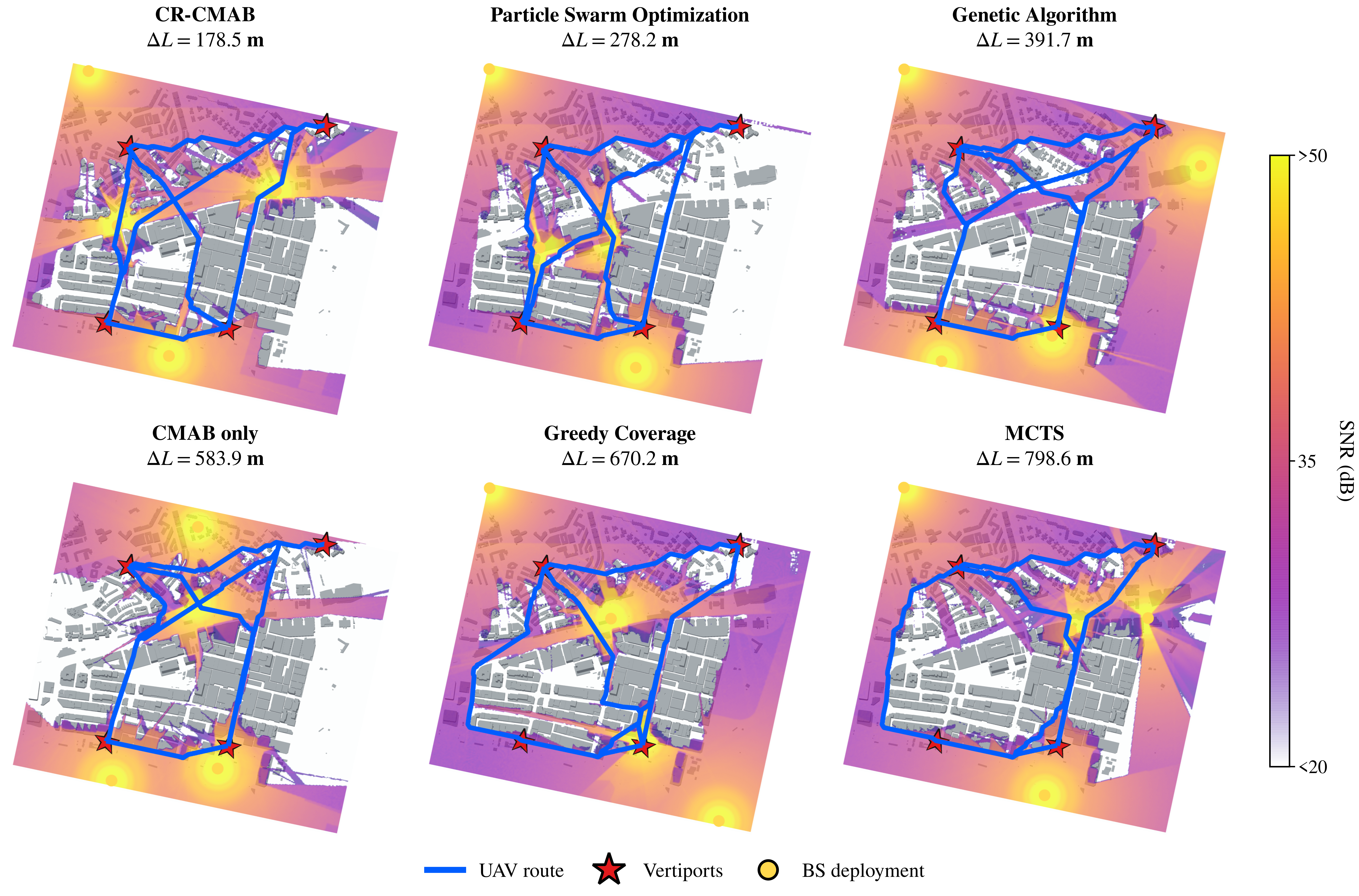} 
    \caption{Visualization of BS deployment and flight corridor planning results for different algorithms when $K=4$.}
    \label{fig:bs_4}
\end{figure*}

The visualizations in Fig.~\ref{fig:bs_4} illustrate the overall concept. The objective is the joint optimization of BS locations and flight corridors to serve a UAV delivery network with four vertiports and six origin-destination pairs in the Kwun Tong urban area shown in Fig.~\ref{fig:landsd_archviz}. In each subfigure, red stars denote the fixed vertiport locations, yellow dots denote the BS locations at 10 m above ground level, blue lines denote the UAV flight corridors at 50 m above ground level, and the colormap shows the signal intensity on the 50 m altitude plane, all overlaid on the Kwun Tong 3D urban model.

In Fig.~\ref{fig:bs_4}, we fix the number of vertiports at $K=4$. Each subfigure shows the BS locations and flight corridor configurations generated by different approaches, with the total path length offset $\Delta L$ displayed in the title. A smaller $\Delta L$ indicates better operational efficiency, as the UAV flight distance is shorter while serving the identical logistics network. We can observe that the planning scheme produced by our proposed CR-CMAB method achieves the smallest $\Delta L$ among all six methods, with a path length offset of just 178.5 m above the theoretical limit of 6277.9 m. This is because CR-CMAB places BSs more effectively in the urban space, allowing wireless signals to penetrate and connect the narrow regions between dense building blocks. As a result, the planned flight corridor can pass through the congested urban area with fewer unnecessary detours. In contrast, Greedy Coverage and MCTS produce much more circuitous flight paths. The BSs selected by these two methods do not sufficiently support the short corridors between vertiports, even though they can still provide feasible communication coverage.

\begin{figure}[h!]  
    \centering
    \includegraphics[width=0.475\textwidth]{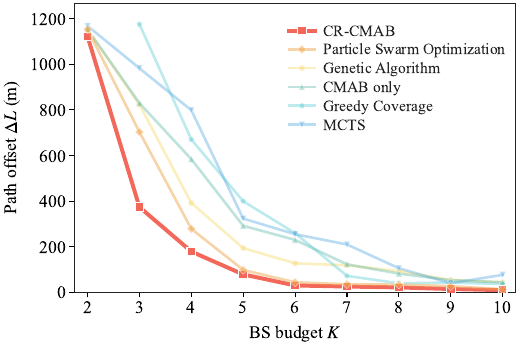}
    \caption{Path length offset of different algorithms under varying BS budgets.}
    \label{fig:overall_performance}
\end{figure}

Fig.~\ref{fig:overall_performance} compares the path length offset of different algorithms under varying BS budgets. As $K$ increases, the path offsets of all methods decrease rapidly at first and then gradually flatten. CR-CMAB achieves the shortest flight corridor path under all tested BS budgets. Its advantage is most evident when $K$ ranges from 3 to 5, where BS resources are still limited and the deployment decision has a strong impact on corridor connectivity. This performance pattern reflects the effectiveness of CR-CMAB under constrained infrastructure resources. In contrast, the CMAB-only result exhibits a substantially larger corridor length than CR-CMAB, demonstrating that reciprocity-driven space expansion brings a clear performance gain to the search algorithm.

\begin{table*}[t]
\centering
\normalsize
\caption{Path length offset and runtime comparison under different BS budgets. Bold values indicate the best results.}
\label{tab:path_runtime_comparison}
\renewcommand{\arraystretch}{1.15}
\setlength{\tabcolsep}{3.2pt}
\begin{tabular}{lcccccccccc}
\hline
\multicolumn{11}{c}{\textbf{Path Length Offset under Different BS Budgets (m)}} \\
\hline
\textbf{Algorithm}
& $K=2$ & $K=3$ & $K=4$ & $K=5$ & $K=6$ & $K=7$ & $K=8$ & $K=9$ & $K=10$ & \textbf{Total} \\
\hline
CR-CMAB 
& \textbf{1122.3} & \textbf{373.7} & \textbf{178.5} & \textbf{78.2} & \textbf{29.7} & \textbf{25.0} & \textbf{21.1} & \textbf{14.0} & \textbf{9.2} & \textbf{1851.7} \\
PSO 
& 1153.4 & 703.0 & 278.2 & 97.8 & 43.8 & 34.9 & 33.4 & 26.6 & 12.7 & 2383.8 \\
GA 
& 1153.4 & 823.2 & 391.7 & 193.6 & 126.8 & 118.3 & 92.7 & 53.9 & 43.6 & 2997.2 \\
CMAB only 
& 1153.4 & 829.6 & 583.9 & 291.5 & 229.7 & 122.6 & 81.6 & 52.8 & 43.9 & 3389.0 \\
Greedy Coverage 
& -- & 1175.8 & 670.2 & 399.4 & 259.0 & 71.8 & 39.1 & 42.6 & 36.2 & 2694.1\rlap{$^\dagger$} \\
MCTS 
& 1168.5 & 982.8 & 798.6 & 323.1 & 254.6 & 209.6 & 105.2 & 39.7 & 76.2 & 3958.3 \\
\hline
\multicolumn{11}{c}{\textbf{Runtime under Different BS Budgets (s)}} \\
\hline
\textbf{Algorithm}
& $K=2$ & $K=3$ & $K=4$ & $K=5$ & $K=6$ & $K=7$ & $K=8$ & $K=9$ & $K=10$ & \textbf{Total} \\
\hline
CR-CMAB 
& 254.3 & 158.4 & 320.4 & 529.3 & 580.1 & 647.2 & 719.1 & 769.8 & \textbf{813.9} & 4792.5 \\
PSO 
& \textbf{121.5} & 457.2 & 598.4 & 1035.1 & 1056.7 & 1386.7 & 1670.4 & 1844.0 & 2096.9 & 10266.9 \\
GA 
& 251.4 & 490.5 & 626.3 & 559.0 & 710.0 & 892.9 & 978.6 & 1034.4 & 1057.7 & 6600.8 \\
CMAB only 
& 472.8 & \textbf{153.7} & \textbf{222.8} & \textbf{317.7} & \textbf{353.0} & \textbf{321.3} & 612.5 & \textbf{761.9} & 870.5 & \textbf{4086.2} \\
Greedy Coverage 
& -- & 350.8 & 304.2 & 354.4 & 411.1 & 463.0 & \textbf{605.3} & 781.3 & 1003.8 & 4273.9\rlap{$^\dagger$} \\
MCTS 
& 619.9 & 313.0 & 471.8 & 744.8 & 1149.1 & 1451.6 & 1567.4 & 1969.6 & 2343.0 & 10630.2 \\
\hline
\multicolumn{11}{l}{$^\dagger$Total is computed over feasible cases only, excluding the infeasible $K=2$ case.} \\
\end{tabular}
\end{table*}

Table~\ref{tab:path_runtime_comparison} provides the detailed path length offset and runtime results for all algorithms. Relative to CMAB only, the fastest method in terms of total runtime, CR-CMAB requires $17.3\%$ more time but reduces the total path length offset by $45.4\%$, from $3389.0$ m to $1851.7$ m. Greedy Coverage has the second shortest total runtime, but fails to find a feasible corridor when $K=2$. PSO, GA, and MCTS require substantially longer runtimes than CR-CMAB, while their path offsets remain larger. Overall, CR-CMAB achieves the best planning performance while keeping the additional computational cost moderate.

\subsection{Ablation Analysis of the Selection Score}

To further examine the internal design of the CR-CMAB algorithm, we conduct an ablation study on the CMAB selection score. Different from the CMAB-only baseline, which removes the reciprocity-driven candidate expansion, this experiment keeps the complete CR-CMAB framework enabled and only changes the composition of the CMAB scoring function. In this way, the effect of each scoring component can be evaluated within the full reciprocity-guided search process.

As introduced in Eq.~\eqref{eq:cmab_score}, the CMAB score used to select each candidate BS comprises three terms: $\bar{r}_i$, $u_i$, and $g_i$. The reward term $\bar{r}_i$ exploits candidates that have appeared in high-quality BS combinations. The confidence term $u_i$ encourages exploration of candidates that have not been sufficiently evaluated, while the coverage term $\lambda_c g_i$ provides spatial guidance by favoring candidates that cover currently uncovered waypoints.

We construct score variants by enabling all non-empty subsets of the reward, confidence, and coverage terms, including single-term, two-term, and full-score configurations. All variants are evaluated under the same reciprocity-driven expansion mechanism, path planning module, and experimental settings. We select $K=3$, $K=4$, and $K=5$ as representative BS budgets, since this is the range where the deployment decision has the strongest impact on system efficiency under a budget constraint.

\begin{table}[h!]
\centering
\caption{Ablation study of scoring components in CR-CMAB. Bold values indicate the best result in each column.}
\label{tab:score_ablation}
\normalsize
\renewcommand{\arraystretch}{1.15}
\begin{tabular}{lccc}
\hline
\multirow{2}{*}{Score Variant} 
& \multicolumn{3}{c}{\textbf{Path Offset $\Delta L$ (m)}} \\
\cline{2-4}
& $K=3$ & $K=4$ & $K=5$ \\
\hline
Reward only 
& 929.0 & 707.9 & 700.1 \\

Confidence only 
& 823.0 & 278.5 & 292.1 \\

Coverage only 
& 679.4 & 179.7 & 92.8 \\

Reward + Confidence 
& 470.3 & 368.0 & 270.2 \\

Reward + Coverage 
& 881.3 & 716.9 & 640.9 \\

Confidence + Coverage 
& 380.0 & 186.6 & 92.5 \\

Full score 
& \textbf{373.7} & \textbf{178.5} & \textbf{78.2} \\
\hline
\end{tabular}
\end{table}

Table~\ref{tab:score_ablation} reports the path offset obtained by different score variants. The full score consistently achieves the smallest offset for all tested BS budgets. Among the reduced variants, coverage-related variants generally perform better than variants without the coverage term, showing that corridor-aware waypoint coverage provides an effective spatial prior for BS selection. The confidence plus coverage variant achieves the closest performance to the full score, which suggests that exploration and coverage complementarity are the main drivers of the search. However, the full score still obtains the best results because the reward term further exploits candidates with strong historical performance. These results confirm that the proposed CR-CMAB benefits from jointly balancing historical feedback, exploration, and heuristic guidance in the BS selection process.

\subsection{Sensitivity to SNR Threshold}

\begin{figure*}[t]
    \centering
    \includegraphics[width=0.75\textwidth]{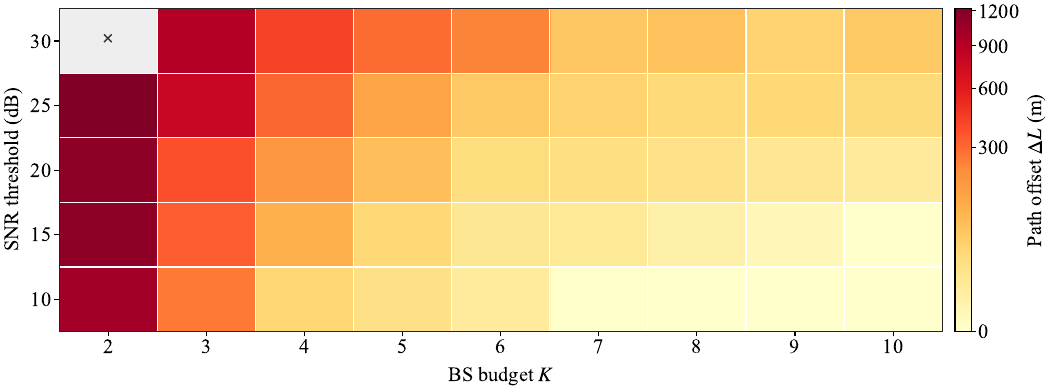}
    \caption{Sensitivity of path offset to BS budget and SNR threshold. The gray cell indicates an infeasible case.}
    \label{fig:snr_threshold}
\end{figure*}

This subsection evaluates the sensitivity of CR-CMAB planning result to the SNR threshold. The tested thresholds represent different levels of link-quality requirements for UAV services. In wireless design guidelines, low-order modulation and low-rate links can operate at relatively low SNR levels, around 10 dB, whereas high-rate or real-time services often require around 25 dB or higher \cite{ciscoMeshSNR}. Similar SNR ranges have also been used in UAV relay communication studies \cite{sliti2026line}. Therefore, in this experiment, lower thresholds can be interpreted as relaxed communication conditions for telemetry, periodic status reporting, or non-real-time monitoring, while higher thresholds represent stricter conditions for video-assisted control, high-definition inspection, and real-time surveillance. By varying the SNR threshold, we examine how different communication service levels affect the feasibility and path efficiency of the planned corridors.

Fig.~\ref{fig:snr_threshold} illustrates the resulting path offset under different SNR thresholds. A clear trend can be observed: for a fixed BS budget, the path offset generally increases as the SNR threshold becomes stricter. This is because a higher threshold shrinks the communication-feasible region and forces the planner to avoid more weakly covered areas. The heatmap also reveals that the most challenging cases occur when the BS budget is small and the SNR requirement is high. In particular, the case with $K=2$ and a 30 dB threshold becomes infeasible. By contrast, when the BS budget is sufficiently large, the path offset remains comparatively small across different thresholds, indicating more stable corridor configuration under varying communication requirements. Overall, stricter communication requirements inevitably constrain the routing space and increase the resulting corridor length.

\subsection{BS Budget Tradeoff and Optimal Selection}

\begin{figure*}[t] 
    \centering
    \includegraphics[width=0.925\textwidth]{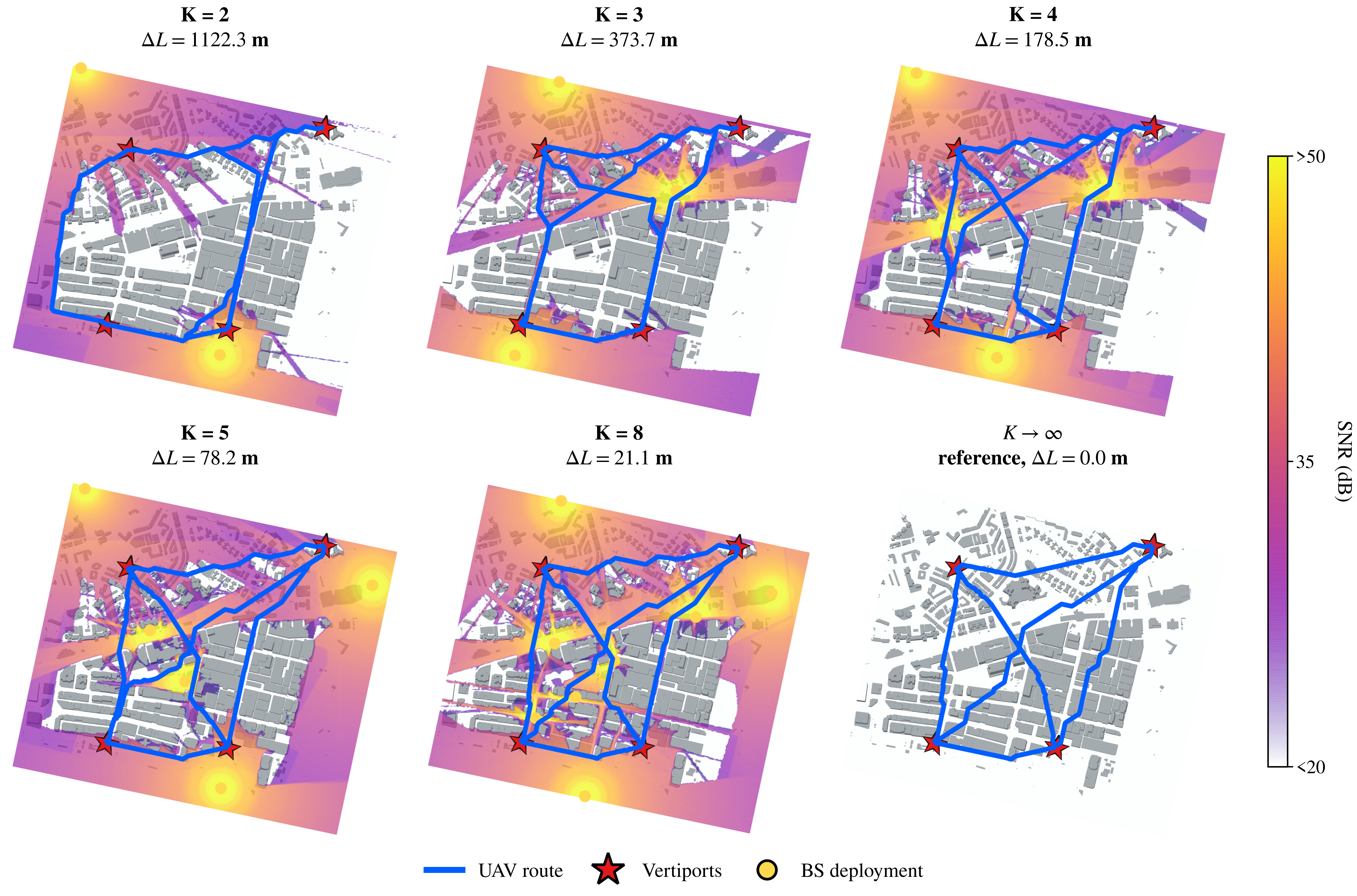} 
    \caption{Visualization of BS deployment and flight corridor planning results for CR-CMAB under varying BS budgets}
    \label{fig:diff_bs}
\end{figure*}

Fig.~\ref{fig:diff_bs} visualizes the results generated by CR-CMAB under different BS budgets. The bottom-right subfigure, labeled $+\infty$ BS, shows the corridor configuration at the theoretical shortest path length $\pi_{\mathrm{ref}}$. When only 2 BSs are deployed, the planned corridors contain a pronounced detour, since the available communication coverage is highly insufficient to support the most direct corridors. As the number of deployed BSs increases, the corridor configuration becomes progressively closer to the reference shortest configuration. The improvement is especially pronounced when the BS budget is increased from $K=2$ to $K=5$, where more regions are connected and the path offset is substantially reduced. At $K=8$, the planned corridor is already very close to the reference path, with only a small residual offset.

Although increasing the BS budget improves corridor connectivity and reduces the path offset, thereby lowering UAV operational cost, it also increases infrastructure investment cost. Therefore, path length alone does not determine which deployment scale should be chosen in practice. To capture this tradeoff in economic terms, we monetize both BS deployment and UAV path offset costs, and select the BS budget that minimizes the total cost.

We consider a five-year evaluation horizon in the following cost analysis. For the infrastructure side, we model the cost of one BS as
\begin{equation}
\label{eq:bs_five_year_cost}
C_{\mathrm{BS}}^{(5)}
=
C_{\mathrm{CAPEX}}
+
5C_{\mathrm{OPEX}}.
\end{equation}
where $C_{\mathrm{CAPEX}}$ denotes the capital expenditure required to deploy one BS site, including the radio equipment and site deployment cost. $C_{\mathrm{OPEX}}$ denotes the annual recurring expenditure required to operate and maintain one BS. We use the dense-city small-cell deployment estimate reported in the ITU 5G cost study, where the CAPEX per site is $54.1$ kUSD \cite{samon2018setting}. Therefore,
\begin{equation}
C_{\mathrm{CAPEX}}=54{,}100~\mathrm{USD/BS}.
\end{equation}

Then, we follow the 5G-Blueprint techno-economic analysis, which estimates the annual maintenance cost as $10\%$ of CAPEX\cite{blueprint2023d33}. Thus,
\begin{equation}
C_{\mathrm{OPEX}}=0.1C_{\mathrm{CAPEX}}=5{,}410~\mathrm{USD/year/BS}.
\end{equation}

For the UAV operation side, we consider a cargo UAV scenario based on the A2Z RDSX Pelican delivery UAV. Its reported operating cost is converted to $1.3\times10^{-4}~\mathrm{USD/(kg\cdot m)}$ \cite{a2zpelican2023}. Assuming a payload of $4$ kg, the distance-based UAV operating cost is
\begin{equation}
c_{\mathrm{uav}}
=1.3\times10^{-4}\times4
=5.2\times10^{-4}~\mathrm{USD/m}.
\end{equation}

\begin{figure*}[h!]
    \centering
    \includegraphics[width=0.85\textwidth]{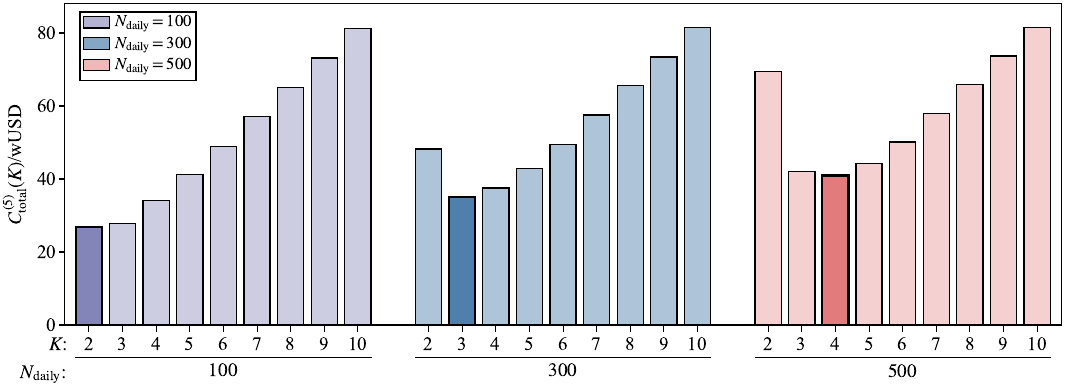} 
    \caption{Five-year cost under different BS budgets and traffic intensities. The highlighted bars indicate the optimal BS budget for each traffic level.}
    \label{fig:best_bs}
\end{figure*}

We use the path offset instead of the total path length to quantify the additional UAV operation cost caused by communication-driven detours. We define $N_{\mathrm{daily}}$ as the number of times that the complete Kwun Tong six-corridor network is operated per day, assuming that all corridors have the same flight intensity. Given the path offset $\Delta L_K$ under BS budget $K$, the five-year extra UAV operating cost is
\begin{equation}
\label{eq:path_cost_five_year}
C_{\mathrm{path}}^{(5)}(K)
=
5\times365\times N_{\mathrm{daily}}
\times
\Delta L_K
\times
c_{\mathrm{uav}},
\end{equation}
where $\Delta L_K$ is measured in meters. The total cost is then
\begin{equation}
\label{eq:total_cost_five_year}
C_{\mathrm{total}}^{(5)}(K)
=
K C_{\mathrm{BS}}^{(5)}
+
C_{\mathrm{path}}^{(5)}(K).
\end{equation}

With the above parameters, Eq. \eqref{eq:total_cost_five_year} can be written as
\begin{equation}
\label{eq:total_cost_simplified}
C_{\mathrm{total}}^{(5)}(K)
=
81{,}150K
+
0.949N_{\mathrm{daily}}\Delta L_K.
\end{equation}

We evaluate three representative traffic intensities: $N_{\mathrm{daily}}=100$, $300$, and $500$, corresponding to low, medium, and high traffic demand, respectively. Fig.~\ref{fig:best_bs} shows the total cost under different BS budgets. For the low-traffic case, the minimum cost is achieved at $K=2$, since the accumulated detour cost is not enough to compensate for additional BS placements. As the traffic intensity increases, the path offset cost becomes more significant, and the optimal budget shifts to $K=3$ for $N_{\mathrm{daily}}=300$ and $K=4$ for $N_{\mathrm{daily}}=500$.

\begin{table}[h!]
\centering
\caption{BS budget selection rule for the six-corridor operation group.}
\label{tab:traffic_budget_rule}
\normalsize
\renewcommand{\arraystretch}{1.15}
\begin{tabular}{cc}
\hline
Daily network operations $N_{\mathrm{daily}}$ & Best BS budget $K^\star$ \\
\hline
$0$--$114$ & 2 \\
$115$--$438$ & 3 \\
$439$--$852$ & 4 \\
$853$--$1763$ & 5 \\
$1764$--$16339$ & 6 \\
\hline
\end{tabular}
\end{table}

The resulting budget rule can be interpreted from a marginal-cost perspective. Adding the $(K+1)$-th BS is economically beneficial only when the five-year path-cost reduction exceeds the five-year cost of one BS:
\begin{equation}
5\times365\times N_{\mathrm{daily}}
\times
\left(\Delta L_K-\Delta L_{K+1}\right)
\times
c_{\mathrm{uav}}
>
C_{\mathrm{BS}}^{(5)}.
\end{equation}
which condition yields the budget selection rule in Table~\ref{tab:traffic_budget_rule}. As $N_{\mathrm{daily}}$ increases, the accumulated cost of detours grows, making additional BS deployment economically justified.

\section{Conclusion}

This paper addressed the joint optimization of BS deployment and flight corridors for a UAV logistics network in complex urban environments. This topic is vital for reducing the operational cost of UAM by balancing infrastructure investment and flight expenditure. Through a channel reciprocity-guided combinatorial multi-armed bandit framework, this research enables efficient BS deployment and UAV path planning under communication constraints.

The main contribution of the CR-CMAB method can be summarized into three aspects. First, the selection score combines exploitation, exploration, and corridor-aware coverage to help the algorithm identify BS combinations with both strong historical feedback and spatial complementarity. Second, the channel reciprocity expansion mechanism utilizes the reversibility of wireless propagation to expand the candidate BS pool with additional promising locations. Third, experimental results validate the proposed framework, showing that CR-CMAB consistently achieves shorter corridors under communication constraints than the baseline methods while maintaining competitive computational efficiency. The cost analysis further demonstrates that the preferred BS deployment scale depends on the expected operation intensity of the UAV delivery network. Future work will incorporate real-world channel measurements to further validate and calibrate the radio maps generated by ray tracing. In addition, the framework can be extended to high-density UAV operations, where conflict avoidance, multi-UAV scheduling, and wireless channel resource allocation become important factors that need to be modeled in future studies.

\section*{Acknowledgment}

This section will be added at a later stage.


\bibliographystyle{IEEEtran}
\bibliography{sample}

\end{document}